\documentclass[letterpaper]{article} 
\usepackage{aaai2026}  
\usepackage{times}  
\usepackage{helvet}  
\usepackage{courier}  
\usepackage[hyphens]{url}  
\usepackage{graphicx} 
\usepackage{natbib}  
\usepackage{caption} 
\usepackage{algorithm}
\usepackage{algorithmic}
\usepackage{newfloat}
\usepackage{listings}
\DeclareCaptionStyle{ruled}{labelfont=normalfont,labelsep=colon,strut=off} 
\floatstyle{ruled}
\newfloat{listing}{tb}{lst}{}
\floatname{listing}{Listing}
\usepackage{booktabs}
\usepackage{amsmath}
\usepackage{amsfonts}
\usepackage{colortbl}       
\usepackage{multirow}
\usepackage{makecell}
\usepackage{cuted}
\usepackage{amsmath}

\title{SGS-3D: High-Fidelity 3D Instance Segmentation via Reliable Semantic Mask Splitting and Growing}
\author{
    Chaolei Wang \textsuperscript{\rm 1}, 
    Yang Luo \textsuperscript{\rm 1},
    Jing Du\textsuperscript{\rm 2}\equalcontrib,
    Siyu Chen\textsuperscript{\rm 3}\equalcontrib,
    Yiping Chen\textsuperscript{\rm 1}\thanks{Corresponding author.},
    Ting Han\textsuperscript{\rm 1}\footnotemark[\value{footnote}]
}

\affiliations{
    \textsuperscript{\rm 1}Sun Yat-sen University, Zhuhai, China\\
    \textsuperscript{\rm 2}University of Waterloo, Waterloo, Canada\\
    \textsuperscript{\rm 3}Jimei University, Xiamen, China\\

    \{wangchlei5, hant23\}@mail2.sysu.edu.cn, chenyp79@mail.sysu.edu.cn
}

\usepackage{bibentry}

\begin{document}

\maketitle

\begin{figure*}[!t]
  \centering
  \includegraphics[width=0.95\linewidth]{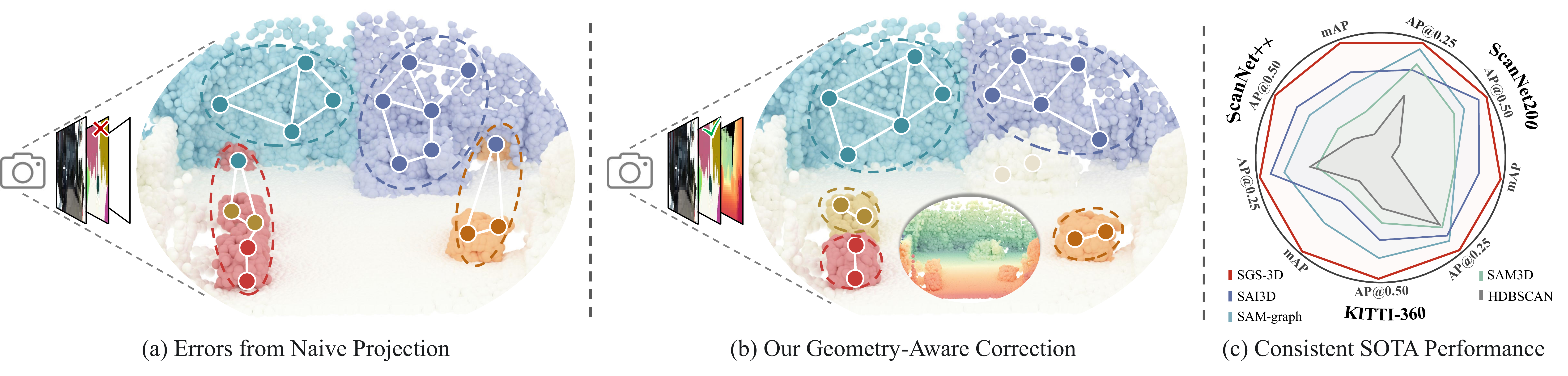}
  \caption{SGS-3D: High-Fidelity segmentation by overcoming ambiguous 2D-to-3D lifting. Previous methods suffer from flawed instance grouping (a), caused by ambiguous 2D semantics and inadequately unhandled occlusions during projection. SGS-3D overcomes this by establishing reliable 3D semantics via occlusion-aware mapping and a novel "split-then-grow" refinement (b). This dual-refinement strategy achieves state-of-the-art accuracy across diverse scenes, especially in challenging, depth-less outdoor environments (c).}
  \label{fig:visualization mapping}
\end{figure*}

\begin{abstract}
Accurate 3D instance segmentation is crucial for high-quality scene understanding in the 3D vision domain. However, 3D instance segmentation based on 2D-to-3D lifting approaches struggle to produce precise instance-level segmentation, due to accumulated errors introduced during the lifting process from ambiguous semantic guidance and insufficient depth constraints. To tackle these challenges, we propose Splitting and Growing reliable Semantic mask for high-fidelity 3D instance segmentation (SGS-3D), a novel "split-then-grow" framework that first purifies and splits ambiguous lifted masks using geometric primitives, and then grows them into complete instances within the scene. Unlike existing approaches that directly rely on raw lifted masks and sacrifice segmentation accuracy, SGS-3D serves as a training-free refinement method that jointly fuses semantic and geometric information, enabling effective cooperation between the two levels of representation. Specifically, for semantic guidance, we introduce a mask filtering strategy that leverages the co-occurrence of 3D geometry primitives to identify and remove ambiguous masks, thereby ensuring more reliable semantic consistency with the 3D object instances. For the geometric refinement, we construct fine-grained object instances by exploiting both spatial continuity and high-level features, particularly in the case of semantic ambiguity between distinct objects. Experimental results on ScanNet200, ScanNet++, and KITTI-360 demonstrate that SGS-3D substantially improves segmentation accuracy and robustness against inaccurate masks from pre-trained models, yielding high-fidelity object instances while maintaining strong generalization across diverse indoor and outdoor environments.
\end{abstract}

\begin{links}
    \link{Code}{https://github.com/wangchaolei7/SGS-3D}
\end{links}

\section{Introduction}

Arbitrary instances segmentation within 3D scenes constitutes a crucial task in various domains, including autonomous driving, virtual reality, and multi-modality scene understanding. While recent methods \cite{felzenszwalb2004efficient, chen2021hierarchical, vu2022softgroup, liang2021instance, engelmann20203d, lu2023query, schult2023mask3d, yu2023transupr, han2024point, luo2025csfnet} yield impressive segmentation results when trained on specifically annotated datasets \cite{dai2017scannet, armeni20163d, robert2023efficient, yeshwanth2023scannet++, liao2022kitti}, their generalization to open-world environments remains limited. In contrast to the inherent difficulties in obtaining extensive 3D annotations, foundation models trained on massive 2D image datasets have shown remarkable performance and strong zero-shot generalization abilities. This observation motivates a promising direction: \textit{utilizing 2D pre-trained foundation models for 3D scene perception and interaction.}

Currently, two predominant paradigms exist for executing the 3D instance segmentation: feature-based methods and mask-based methods. Feature-based methods \cite{peng2023openscene, takmaz2023openmask3d, ding2023pla, hegde2023clip, yang2024regionplc, lee2024segment, huang2024openins3d} primarily focus on learning robust feature representations from 3D point clouds or by lifting features from 2D modalities \cite{oquab2023dinov2}. Instance segmentation is subsequently achieved by classifying, grouping, or decoding these learned features into instance-level masks. However, these methods commonly suffer from inefficient training and error propagation, which arising from ambiguous feature representations stemming from 2D-to-3D lifting or the inherent high-dimensional feature space.

To mitigate these concerns, mask-based methods leverage the semantic masks extracted from input registered images using Segment Anything Model (SAM) \cite{kirillov2023segment} to derive instance-level assignments for each 3D geometric primitive in the 3D scene, consolidating these 3D primitives iteratively based on their projected masks \cite{yang2023sam3d, lu2023ovir, xu2023sampro3d, nguyen2024open3dis, huang2024segment3d}. Despite this advancement, the robust propagation of semantic information from these projected masks is often overlooked, with a primary focus on optimizing strategies for merging 3D geometric primitives while neglecting fine-grained geometric cues. This oversight leads to continuous error accumulation during scene segmentation, ultimately yielding imprecise results, particularly for adjacent but separate instances, as il-lustrated in Figure \ref{fig:visualization mapping}. Although some existing methods incorporate depth information from sensors to enhance back-projection accuracy \cite{guo2024sam, yin2024sai3d, zhao2025sam2object}, this reliance is problematic for two key reasons: depth sensors inherently struggle on textureless and highly reflective surfaces, and moreover, they are often entirely unavailable in wild scenes, severely compromising segmentation performance.

To confront these issues, we propose SGS-3D, a training-free refinement and segmentation framework that jointly fuse semantic and geometric information to accurately register 2D semantic instance masks to 3D geometric primitives. Inspired by prior work, we introduce 3D geometric over-segmentation to decompose objects within the scene into primitives, which serve as the basic units for instance segmentation. Moreover, we reveal the significance of precise depth constraints, which serve as a powerful tool for propagating reliable and fine-grained semantic cues to 3D point clouds. Specifically, during the mask refinement phase, we first effectively establish a point-to-pixel mapping, without relying on true depth maps and without sacrificing precision. Leveraging this mapping, the co-occurrence of 3D geometric primitives effectively prevents the accumulation of ambiguous semantic errors. In the geometry refinement phase, we split 3D semantic masks within the density space to further refine semantic guidance and maintain semantic quality. After geometric optimization, our method consolidates the refined 3D semantic masks with geometric primitives by performing feature similarity matching, which enhances the completeness of object instances while reducing redundancy. To summarize our contributions in a few words:

\begin{itemize}
   \item We propose SGS-3D, a novel training-free framework that achieves high-fidelity 3D instance segmentation through a principled "split-then-grow" strategy, jointly leveraging semantic and geometric cues to overcome error accumulation in existing 2D-to-3D lifting approaches.
   \item We introduce an occlusion-aware point-image mapping that ensures accurate mask-to-point correspondence without ground-truth depth, and develop a visibility-based co-occurrence filtering that effectively removes ambiguous 2D masks while achieving 4$\times$ computational speedup and robust cross-scene generalization.
   \item We design a semantic-guided aggregation pipeline featuring density-based spatial splitting to generate pure semantic-geometric seeds, followed by feature-guided growing that intelligently aggregates fragmented instances into complete objects, particularly excelling in challenging interwoven scenarios.
\end{itemize}

\begin{figure*}[!t]
  \centering
  \includegraphics[width=0.95\textwidth]{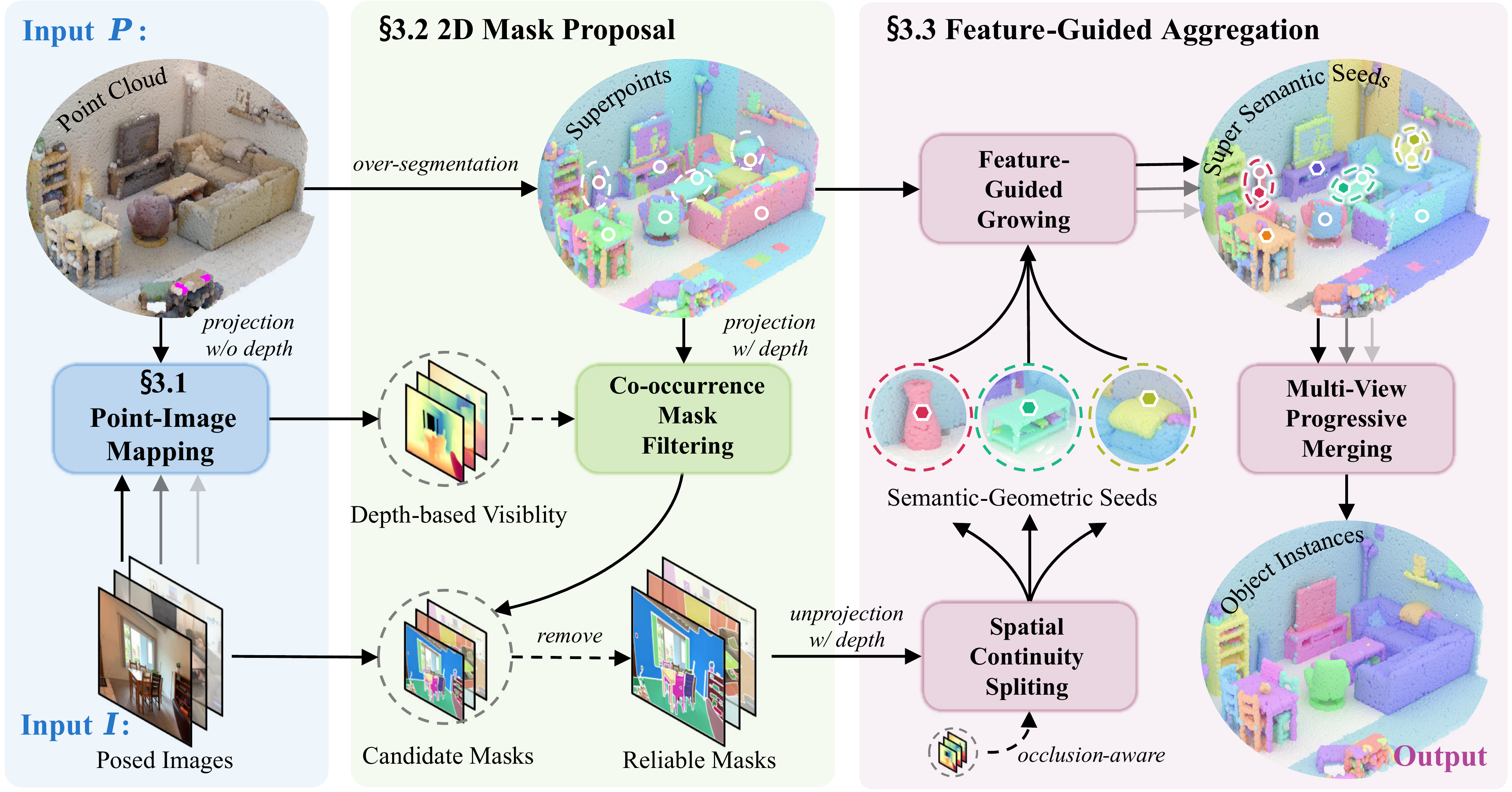}
  \caption{Overview of the training-free SGS-3D pipeline. Our method begins by computing robust, occlusion-aware point-image mappings without requiring ground-truth depth (§ 3.1). In the 2D Mask Proposal stage (§ 3.2), these mappings guide a Co-occurrence Mask Filtering process to prune ambiguous candidate masks, yielding a set of reliable 2D masks. These are then lifted to 3D and fed into the Feature-Guided Aggregation stage (§ 3.3), which first uses Spatial Continuity Splitting to generate pure semantic-geometric seeds. Subsequently, Feature-Guided Growing expands these seeds into complete instances, which are finally consolidated via Multi-View Progressive Merging to produce the final, high-fidelity object instances.}
  \label{fig: pipeline}
\end{figure*}


\section{Methodology}

\subsection{Method Overview and Preliminaries}
The proposed SGS-3D comprises three main components: (1) Point-Image Mapping, (2) 2D Mask Proposal Module, and (3) Semantic-Guided Aggregation, as illustrated in Figure~\ref{fig: pipeline}. Our primary objective is to generate a set of high-quality, class-agnostic 3D instance proposals by leveraging semantic and spatial cues from multiple 2D views. Our method takes as input a 3D point cloud $P = \{p_i \in \mathbb{R}^6\}_{i=1}^N$ (XYZRGB) and a set of $T$ registered images $\{I_t\}_{t=1}^T$ with their corresponding camera parameters, including intrinsics $\mathbf{K}_t$ and pose $\mathbf{T}_t$.

We first over-segment the point cloud $P$ into a set of $U$ superpoints, $\mathcal{S} = \{s_u\}_{u=1}^U$, which serve as our fundamental processing primitives. These spatially coherent and geometrically compact groups of points improve computational efficiency and help maintain structural consistency.
To adapt to diverse 3D scenarios, we employ method in \cite{dai2017scannet} for indoor mesh inputs and \cite{robert2023efficient} for outdoor unstructured point clouds. Furthermore, for each superpoint $s_u$, we assume access to a pre-computed feature vector $\mathbf{f}_u \in \mathbb{R}^D$, derived from a pre-trained segmentation model \cite{schult2023mask3d}, which captures its high-dimensional feature. 

\subsection{Point-Image Mapping} \label{sec:point-image mapping}

We start by establishing a robust and efficient mapping between the points in $P$ and the images in $I$. A point-image pair $(p, I_t)$ is considered compatible if the point $p$ is visible in the image $I_t$. That is, $p$ lies within the camera's viewing frustum and is not occluded by any other point that is closer to the camera. For each compatible pair, we define its re-projection $\text{pix}(p, I_t)$ as the 2D pixel coordinate in image $I_t$ where the point $p$ is projected. However, computing this mapping is non-trivial, especially in multi-view scenarios where occlusions are common.

\subsubsection{Depth Construction}

In contrast to existing methods that rely on ground-truth depth maps from dedicated sensors, we propose an efficient mapping strategy to compute visibility directly from the point cloud and camera parameters using Z-buffering \cite{ravi2020accelerating}. For a given image $I_t$ with its associated camera parameters $(\mathbf{K}_t, \mathbf{T}_t)$, the entire point cloud $P$ is projected into the camera's image plane. The transformation $\pi$ maps a homogeneous world point $p_i$ to pixel coordinates $(u_i, v_i)$ with depth $z_i$, according to $z_i [u_i, v_i, 1]^T = \mathbf{K}_t \mathbf{T}_t^{-1} \mathbf{p}_i$. To resolve occlusions, a depth buffer $\mathbf{D}_t \in \mathbb{R}^{H \times W}$ is initialized with infinite values. The final value at each pixel $(u,v)$ is populated via a highly parallel rasterization process on the GPU, which efficiently solves for the minimum depth over the set of all points projecting to that location:
\begin{equation}
    \mathbf{D}_t(u,v) = \min \{ z_i \mid p_i \in P, \pi(p_i, \mathbf{K}_t, \mathbf{T}_t) = (u,v) \}.
\label{eq:Z-buffering}
\end{equation}

\begin{figure}[t]
  \centering
  \includegraphics[width=0.45\textwidth]{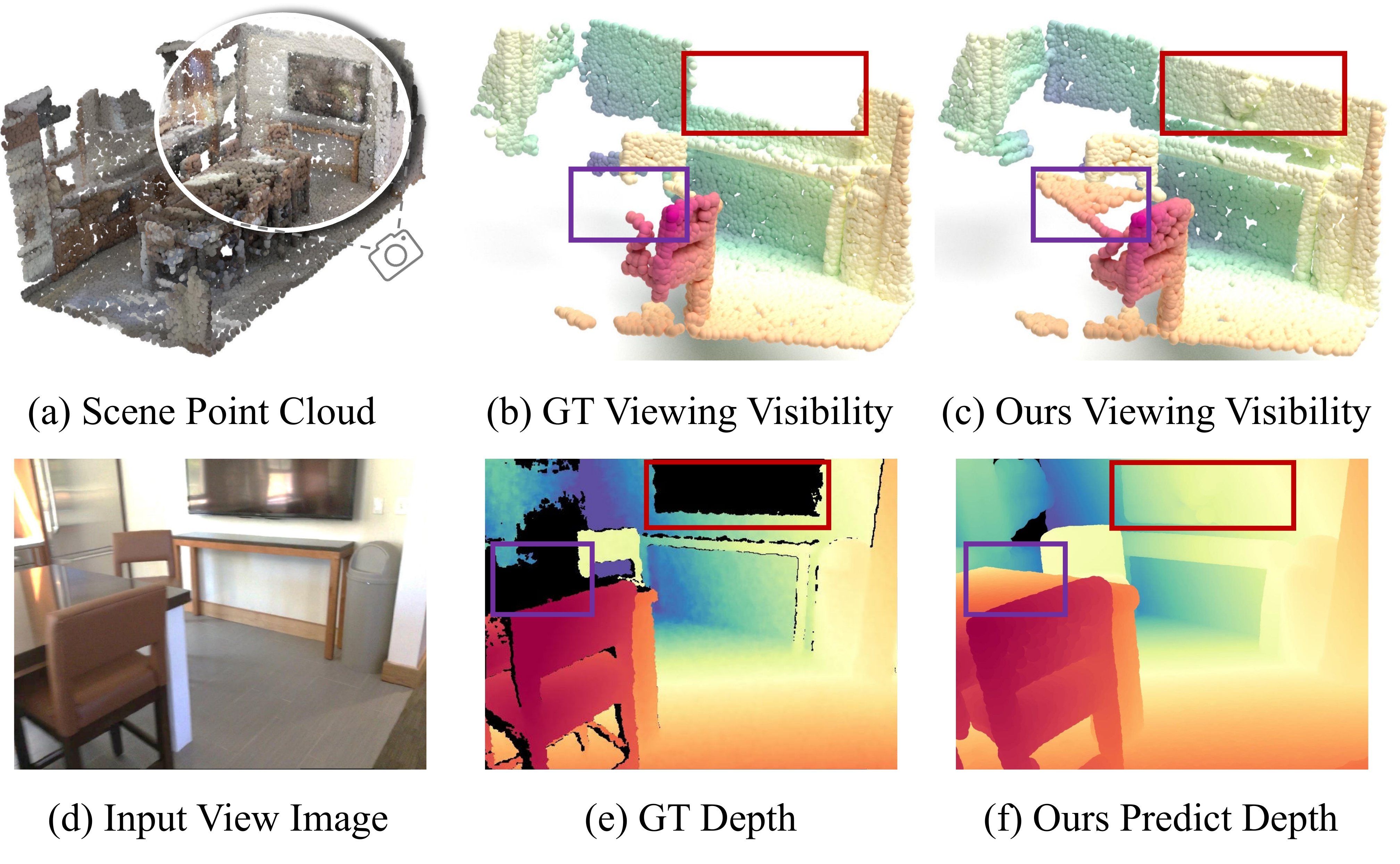}
  \caption{Our method constructs valid visibility mapping for each point. While depth sensors struggle on textureless and highly reflective surfaces, our approach remains effective.}
  \label{fig: depth}
\end{figure}

\subsubsection{Viewing Visibility Verification}
Subsequent to the generation of the metric depth buffer $\mathbf{D}_t$, we finalize the visibility status for each point. The per-point visibility for a view $t$ is encapsulated in a binary tensor $\mathcal{V}_t \in \{0, 1\}^N$, whose elements are computed via a composite frustum and occlusion check:
\begin{align}
\mathcal{V}_t(i) &= \mathbf{1}(0 \le u_i < W \land 0 \le v_i < H) \nonumber \\
&\quad \cdot \mathbf{1}(|z_i - \mathbf{D}_t(u_i, v_i)| \le \tau_{vis}),
\label{eq:mapping}
\end{align}
where $(u_i, v_i)$ and $z_i$ are the projected pixel coordinates and true computed camera-space depth of point $p_i$, respectively, $\mathbf{1}(\cdot)$ is the indicator function, and $\tau_{vis}$ is the depth congruence tolerance. Simultaneously, we maintain a definitive point-to-pixel correspondence map $\mathcal{M}_{pix,t} \in \mathbb{Z}^{N \times 3}$, which stores the valid mapping $[u_i, v_i, 1]^T$ for each point where $\mathcal{V}_t(i)=1$. 
This verification yields the precise, per-view mapping essential for subsequent sections, as illustrated in Figure \ref{fig: depth}.

\subsection{2D Mask Proposal}
\label{sec:mask generation and filtering}

\subsubsection{Mask Generation}
Mask-based 3D instance segmentation involves generating masks for the target objects based on input images. For each image $I_t$ in a scene sequence $I = \{I_t\}_{t=1}^T$, the objective is to produce a corresponding set of binary masks, $\mathcal{M}_{2D,t} = \{M_{t,k}\}_{k=1}^{K_t}$. With the emergence of foundational models like SAM \cite{kirillov2023segment}, mask prediction has become highly precise. However, SAM lacks targeted guidance and often segments objects that are not of interest. Therefore, we apply Grounding-DINO \cite{liu2024grounding} to extract box prompts from the images with text prompts. The high-confidence box prompts among them are then used to generate the initial mask predictions.

\begin{figure}[t]
  \centering
  \includegraphics[width=0.4\textwidth]{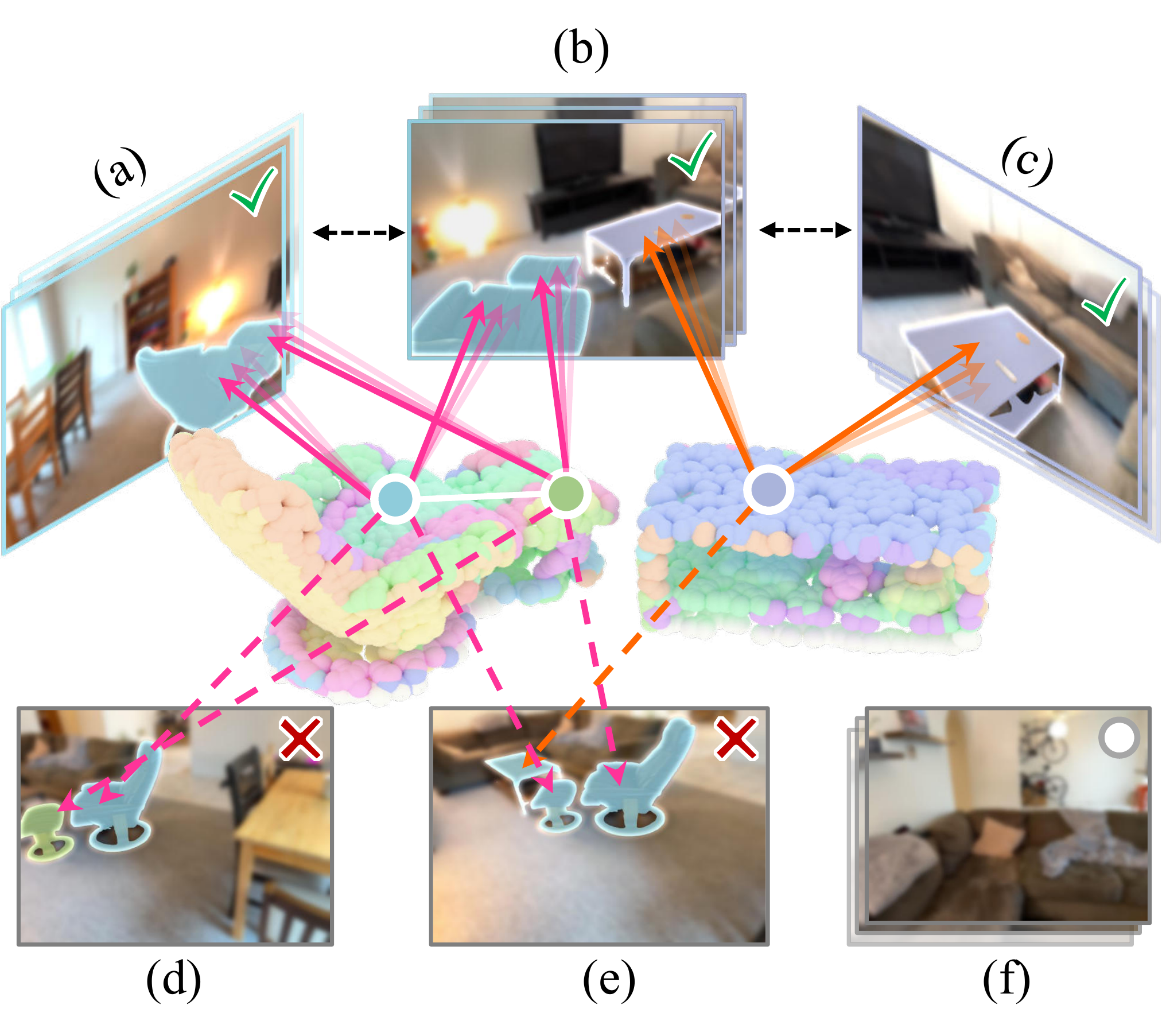}
  \caption{Co-occurrence mask filtering strategy. Co-occurrence scores between superpoints are constructed from accurate 2D mask sets (a-c). Over-segmented (d) and under-segmented (e) masks exhibiting low scores are then removed from the candidate masks list, where image sets with $\mathcal{P}_{vis,m}^j = 0$ (f) are excluded from the calculation.}
  \label{fig:co-occurrence}
\end{figure}

\subsubsection{Co-occurrence Mask Filtering}
Through introducing text prompts via Grounding-DINO, the number of semantically ambiguous masks should progressively decline. However, experiments reveal that although the quantity of major ambiguous masks reduces, some minor ambiguous ones still remain. 

In fact, the binary masks $\mathcal{M}_{2D,t}$ predicted by the trained 2D vision foundation model exhibit discreteness, which often leads to inaccuracies and inconsistencies in object boundary area predictions across different views. This limitation may increase the instability of the semantic guidance of 2D masks. Unlike Maskclustering \cite{yan2024maskclustering}, which purely relies on image pixel consistency between views, we leverage viewing visibility of superpoints to exclusively enhance robustness against inaccurate masks.

Specifically, for superpoints belonging to the same instance, we accumulate their mutual visibility across multiple masks and views in Figure \ref{fig:co-occurrence}. We leverage our pre-computed point-to-pixel mappings $\mathcal{M}_{pix,t} \in \mathbb{Z}^{N \times 3}$ to establish the association between a superpoint and a 2D mask. A point $p_i$ is considered to fall within a mask $M_m$ in view $j$ if it is physically visible ($\mathcal{V}_j(i)=1$) and its projected pixel coordinates $(u_i, v_i)$ lie inside the mask area. Let the set of all superpoints in the scene be $\mathcal{S} = \{s_u\}_{u=1}^U$. We define the association between a mask $M_m$ and the superpoints it makes visible in view $j$, denoted as the set $\mathcal{P}_{vis,m}^j$:

\begin{equation}
\small
\mathcal{P}_{vis,m}^j = \{s_u \in \mathcal{S} \mid
\tfrac{| \{p_i \in s_u \mid \mathcal{V}_j(i)=1 \land M_m(\pi_j(p_i))=1 \} |}%
{|s_u|} > 0.5 \}.
\label{eq:superpoint_visibility_revised}
\end{equation}

Based on this, we define the visibility co-occurrence score $c_{m}$ for each mask, which measures its average consistency with all other masks in the candidate set:
\begin{equation}
\small
 c_{m} = \frac{1}{(K_{2D}-1)T} \sum_{\substack{n=1, n \neq m}}^{K_{2D}} \sum_{j=1}^{T} \frac{|\mathcal{P}_{vis,m}^j \cap \mathcal{P}_{vis,n}^j|}{\sqrt{|\mathcal{P}_{vis,m}^j| \cdot |\mathcal{P}_{vis,n}^j|}},
\label{eq:co-occurrence_filter_superpoint}
\end{equation}
where $\mathcal{P}_{vis,m}^j$ represents the set of visible superpoints for mask $M_m$ in view $j$, $T$ is the total number of views, and $K_{2D}$ is the total number of masks in the candidate set. This principled score allows us to rank masks by their cross-view consistency. To prevent error propagation, masks exhibiting low co-occurrence scores, which indicate inconsistent visibility patterns, are identified as outliers and pruned from the candidate set.

\begin{figure}[t]
    \centering
    \includegraphics[width=0.45\textwidth]{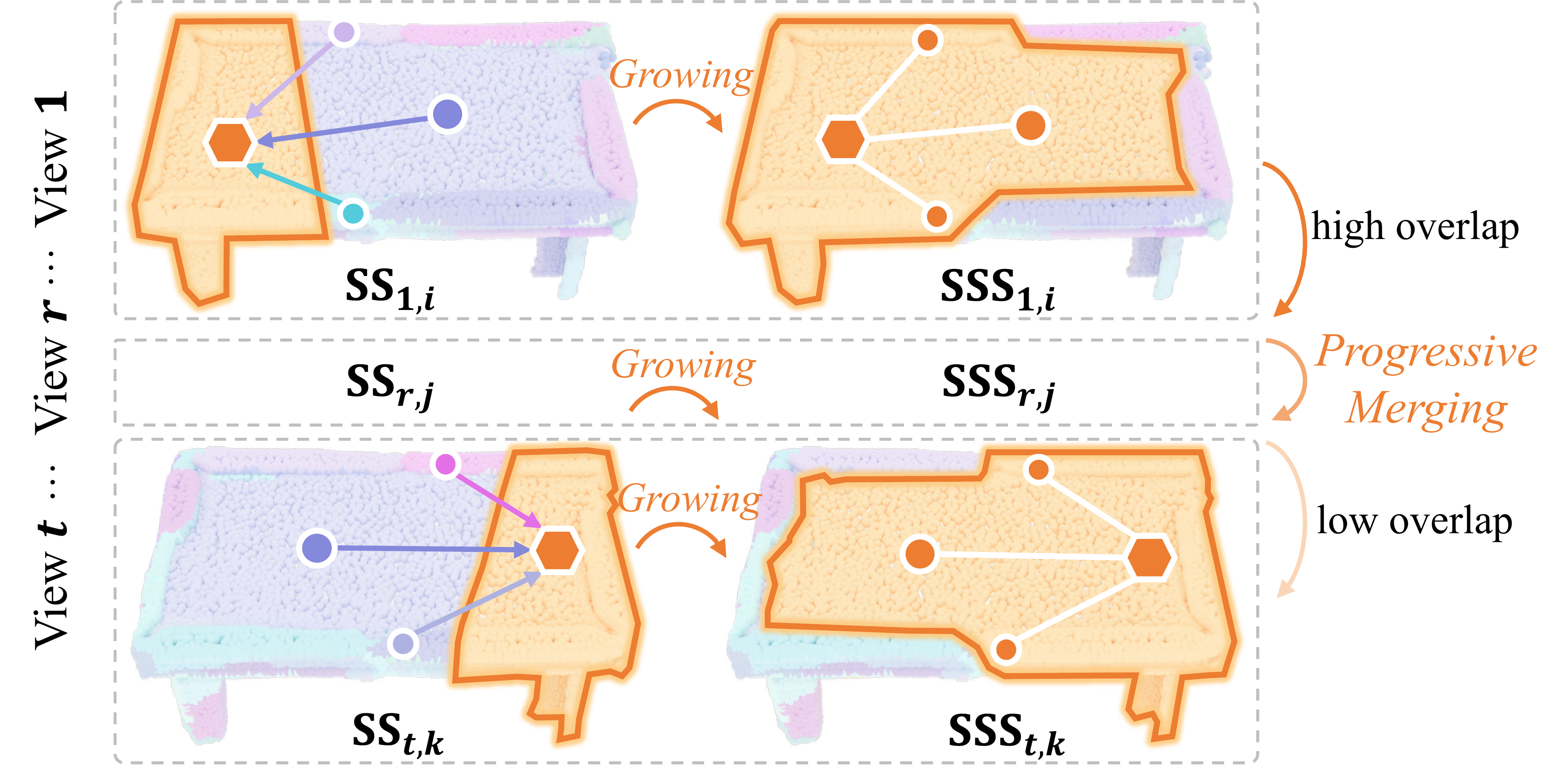}
    \caption{Semantic-guided aggregation. Within a single view, a Semantic-geometric Seed (SS) (orange) is expanded into a Super Semantic Seed (SSS) by merging with neighboring superpoints (colorful). Subsequently, these proposals from other views are progressively merged to form the final object instance.}
    \label{fig: growing}
\end{figure}

\begin{figure*}[!t]
  \centering
  \includegraphics[width=1\textwidth]{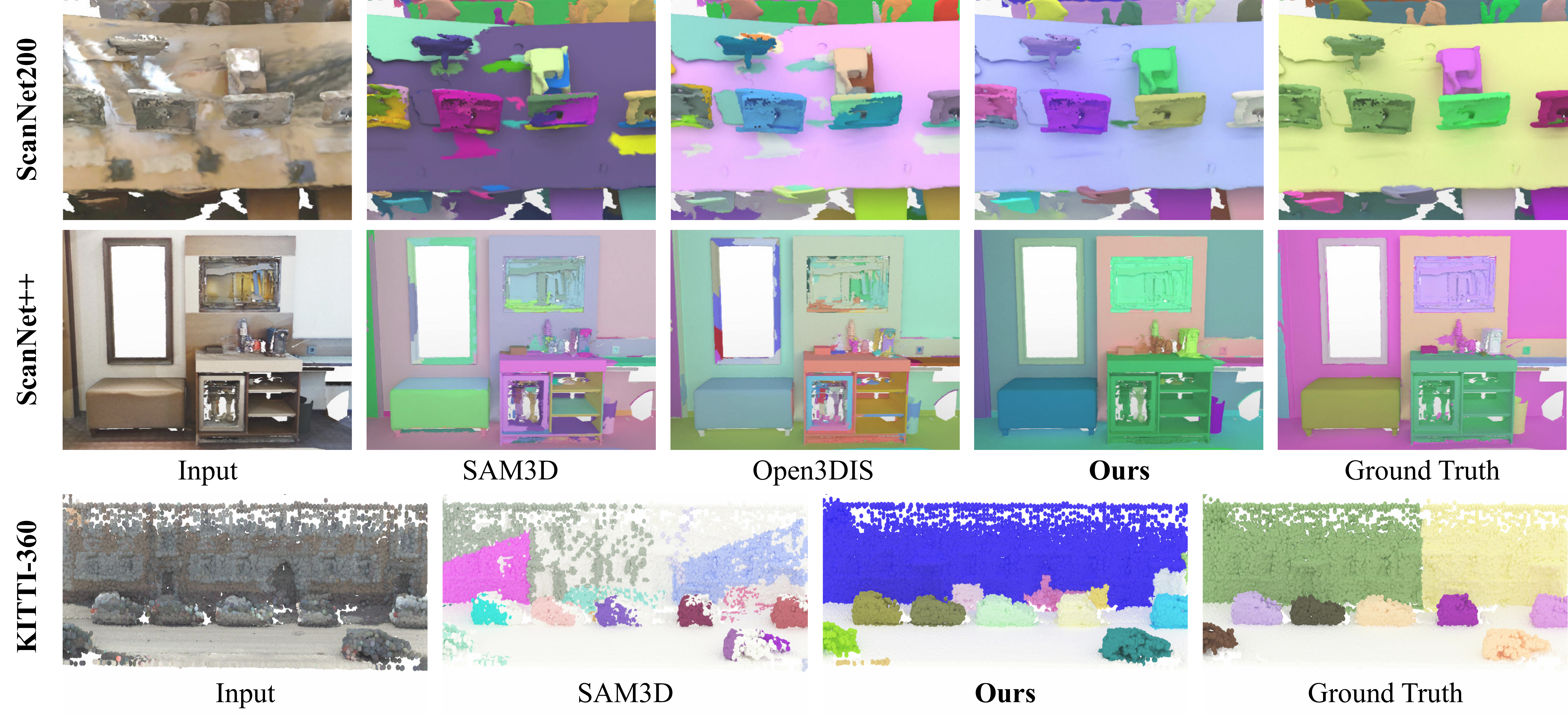}
  \caption{Qualitative comparison on indoor and outdoor datasets. Our method obtains more accurate segmentation results and significantly reduces over-segmentation or fragmented instances compared to previous method.}
  \label{fig:qulitive_scannet_kitti}
\end{figure*}

\subsection{Semantic-Guided Aggregation}
Existing methods often compromise segmentation quality by either directly unprojecting 2D masks onto the point cloud \cite{yang2023sam3d, yan2024maskclustering} or by solely using them as a semantic prior to coarsely guide the iterative merging of superpoints \cite{yin2024sai3d, nguyen2024open3dis}. In such rough methods, the interplay between semantic and geometric information is not deeply exploited and leads to over-segmentation of object instances. To solve this problem, our unique insight is that leveraging reliable 3D semantics can enhance 3D segmentation and improve the robustness of instance aggregation.

\subsubsection{Spatial Continuity Splitting}
Leveraging the occlusion-aware viewing visibility, we lift the filtered 2D masks from each view $t$ onto point cloud. These point sets, aggregated by their shared semantic identity, are termed 3D semantic masks, $\mathcal{M}_{3D,t} = \{M_{t,k}\}_{k=1}^{K_t}$. While the 2D semantic consistency derived from vision models has proven effective in prior work, the intrinsic geometric coherence of the 3D point cloud itself is often overlooked. We experimentally find that objects with similar appearances are erroneously grouped into a single instance, despite being spatially distinct in the point cloud (see supplementary, Figure 10). As a solution, we apply HDBSCAN \cite{campello2013density} on each of the 3D semantic masks ($M_{t,k} \in \mathcal{M}_{3D,t}$) which splits them into new clusters based on spatial contiguity. We refer to these dense clusters as semantic-geometric seeds, forming a new collection of fine-grained masks $\mathcal{M}_{seed,t} = \{M'_{t,j}\}_{j=1}^{K'_t}$. This density-based splitting utilizes geometric properties to refine the semantics of the segmentation results, ensuring a fine-grained distinction between instances and mitigating the propagation of ambiguous semantics. The effectiveness of this geometric refinement in resolving semantic ambiguity is significant, as our ablation study shows it single-handedly surpasses existing state-of-the-art methods (see Table \ref{table:ablation}).

\subsubsection{Single-View Feature-Guided Growing}
While the preceding density-based splitting excels at purifying seeds, producing $\mathcal{M}_{seed,t}$, the challenge is ensembling these over-segmented superpoints into complete instances. To address this, we introduce a feature-guided growing process. We leverage rich pre-trained features \cite{schult2023mask3d} to bridge the spatial gaps between these semantic-geometric seeds and their corresponding object parts.

Rather than applying rigid thresholds, our growing process is guided by a unified affinity score that naturally balances semantic coherence with spatial adjacency, as illustrated in Figure \ref{fig: growing}. For a given seed $M'_{t,j} \in \mathcal{M}_{seed,t}$ and a candidate neighbor superpoint $s_u$, we define this score as:
\begin{equation}
\text{Affinity}(M'_{t,j}, s_u) = \text{sim}(\bar{\mathbf{f}}_{t,j}, \mathbf{f}_u) \cdot \text{Overlap}(M'_{t,j}, s_u),
\label{eq:affinity_score}
\end{equation}
where $\text{sim}(\bar{\mathbf{f}}_{t,j}, \mathbf{f}_u)$ is the cosine similarity between the seed's mean feature and the candidate's mean feature, and $\text{Overlap}(\cdot, \cdot)$ measures their spatial overlap, calculated as the IoU of their 3D point sets. This multiplicative form ensures that a high affinity is only achieved when both semantic similarity and spatial overlap are strong.

\begin{table*}[!t]
  \centering
  \setlength{\tabcolsep}{2pt}
  \begin{tabular}{l|c|c|ccc|ccc|ccc}
    \toprule
    \multirow{2}{*}{Methods} & \multirow{2}{*}{Proc. \& Year} & \multicolumn{1}{c|}{\multirow{2}{*}{VFM}}
    & \multicolumn{3}{c}{ScanNet200}
    & \multicolumn{3}{c}{ScanNet++}
    & \multicolumn{3}{c}{KITTI-360} \\
    \cmidrule(lr){4-6} \cmidrule(lr){7-9} \cmidrule(lr){10-12}

    & & & mAP & AP$_{50}$ & AP$_{25}$ & mAP & AP$_{50}$ & AP$_{25}$ & mAP & AP$_{50}$ & AP$_{25}$ \\
    \midrule
    \multicolumn{12}{l}{training-\textit{dependent}}\\
    UnScene3D & CVPR24 & DINO & 15.9 & 32.2 & 58.5 & - & - & - & - & - & - \\
    Segment3D & ECCV24 & SAM & - & - & - & 12.0 & 22.7 & 37.8 & - & - & - \\
    SAM-graph & ECCV24 & SAM & 22.1 & 41.7 & \underline{62.8} & 15.3 & 27.2 & 44.3 & \underline{23.8} & \underline{37.2} & \underline{49.1} \\
    \midrule
    \multicolumn{12}{l}{training-\textit{free}} \\
    HDBSCAN & ICDMW17 & - & 2.9 & 8.2 & 33.1 & 4.3 & 10.6 & 32.3 & 9.3 & 18.9 & 39.6 \\
    Felzenszwalb & IJCV04 & - & 4.8 & 9.8 & 27.5 & 4.1 & 9.2 & 25.3 & - & - & - \\
    SAM3D & ICCVW23 & SAM & 20.9 & 34.8 & 51.4 & 9.3 & 16.6 & 29.5 & 13.0 & 24.2 & 41.1 \\
    Open3DIS & CVPR24 & GD-SAM & 29.7 & 45.2 & 56.8 & 18.5 & 35.5 & 44.3 & - & - & - \\
    SAI3D & CVPR24 & SAM & 28.2 & \underline{47.2} & 48.5 & 17.1 & 31.1 & 49.5 & 16.5 & 30.2 & 45.6 \\
    SAM2Object & CVPR25 & SAM2 & - & - & - & 20.2 & 34.1 & 48.7 & - & - & - \\
    \textbf{Ours$^{-}$}
    & - 
    & GD-SAM
    & \underline{30.5}  
    & \underline{47.2}  
    & 62.2  
    & \underline{22.5}
    & \underline{36.6} 
    & \underline{53.0} 
    & \textbf{32.9}$\,_\text{{+16.4}}$  
    & \textbf{43.7}$\,_\text{{+13.5}}$ 
    & \textbf{53.4}$\,_\text{{+7.8}}$  \\
    \textbf{Ours$^{+}$}
    & - 
    & GD-SAM
    & \textbf{34.3}$\,_\text{{+4.6}}$  
    & \textbf{51.3}$\,_\text{{+4.1}}$   
    & \textbf{64.6}$\,_\text{{+7.8}}$   
    & \textbf{23.7}$\,_\text{{+3.5}}$  
    & \textbf{39.6}$\,_\text{{+4.1}}$  
    & \textbf{54.3}$\,_\text{{+4.8}}$  
    & N/A 
    & N/A 
    & N/A \\
    \bottomrule
  \end{tabular}
  \caption{Quantitative results compared with conventional clustering methods and diverse 2D-to-3D lifting methods on ScanNet200, ScanNet++ and KITTI-360. "VFM" denotes the vision foundation models. Ours$^{-}$ and Ours$^{+}$ represent our method without and with ground-truth (GT) depth, respectively. N/A indicates results are not applicable as KITTI-360 lacks the required GT depth. All 2D-to-3D lifting baselines use GT depth in indoor datasets. Best results are in \textbf{bold} (\%), and the second-best are \underline{underlined}.} 
  \label{table:quantitative results}
\end{table*}

\begin{table}[htbp]
  \centering
  \setlength{\tabcolsep}{7pt}
  \begin{tabular}{c|cccccc}
    \toprule
    No. & CMF & SCP & FGG & mAP & AP$_{50}$ & AP$_{25}$ \\
    \midrule 
    1 & & & & 25.9 & 41.9 & 57.3 \\
    2 & \checkmark & & & 27.8 & 44.0 & 59.0 \\
    3 & \checkmark & \checkmark & & 30.1 & 46.8 & 61.4 \\
    4 & \checkmark &  & \checkmark & 29.7 & 45.2 & 58.9 \\
    5 & \checkmark & \checkmark & \checkmark & 34.3 & 51.3 & 64.6 \\
    \bottomrule
  \end{tabular}
  \caption{Ablation of the effectiveness of our proposed components. CMF indicates the co-occurrence mask filtering; SCP indicates the spatial continuity Splitting; FGG indicates the feature-guided growing.}
  \label{table:ablation}
\end{table}

At each expansion step, we merge the unassigned neighbor with the highest affinity score into the seed. This iterative process continues until the affinity scores of all remaining neighbors become negligible. By optimizing for the highest affinity, object instances can correctly associate disparate parts and expand their boundaries for the final aggregation stage.

\subsubsection{Multi-View Progressive Merging}
The single-view growing process generates high-fidelity instance proposals, which are essentially the grown semantic-geometric seeds from $\mathcal{M}_{seed,t}$. However, each proposal is inherently limited to the information available from its originating viewpoint. To reconcile these partial views into complete objects, we employ a progressive, multi-view merging strategy. Inspired by \cite{yin2024sai3d}, we initially merge proposals from different views only if their 3D spatial overlap is very high, ensuring only unambiguous matches are made. Subsequent iterations systematically relax this overlap requirement, allowing fragmented parts of the same large instance to be fused. This hierarchical process robustly assembles the final, complete 3D object instances for the entire scene, effectively consolidating information from multiple views.

\section{Experiments}
\subsection{Experimental Setup}

We evaluate our method on three prevalent benchmarks: ScanNet200 \cite{dai2017scannet}, ScanNet++ \cite{yeshwanth2023scannet++} for indoor scenes, and KITTI-360 \cite{liao2022kitti} for outdoor environments. Following established protocols \cite{rozenberszki2024unscene3d, schult2023mask3d, takmaz2023openmask3d}, we compute AP scores at 50\% (AP$_{50}$) and 25\% (AP$_{25}$) intersection-over-union (IoU) thresholds, as well as the mean AP (mAP) averaged over 50\% to 95\% IoU thresholds in 5\% increments.

\begin{figure}[htbp]
    \centering
    \includegraphics[width=0.95\linewidth]{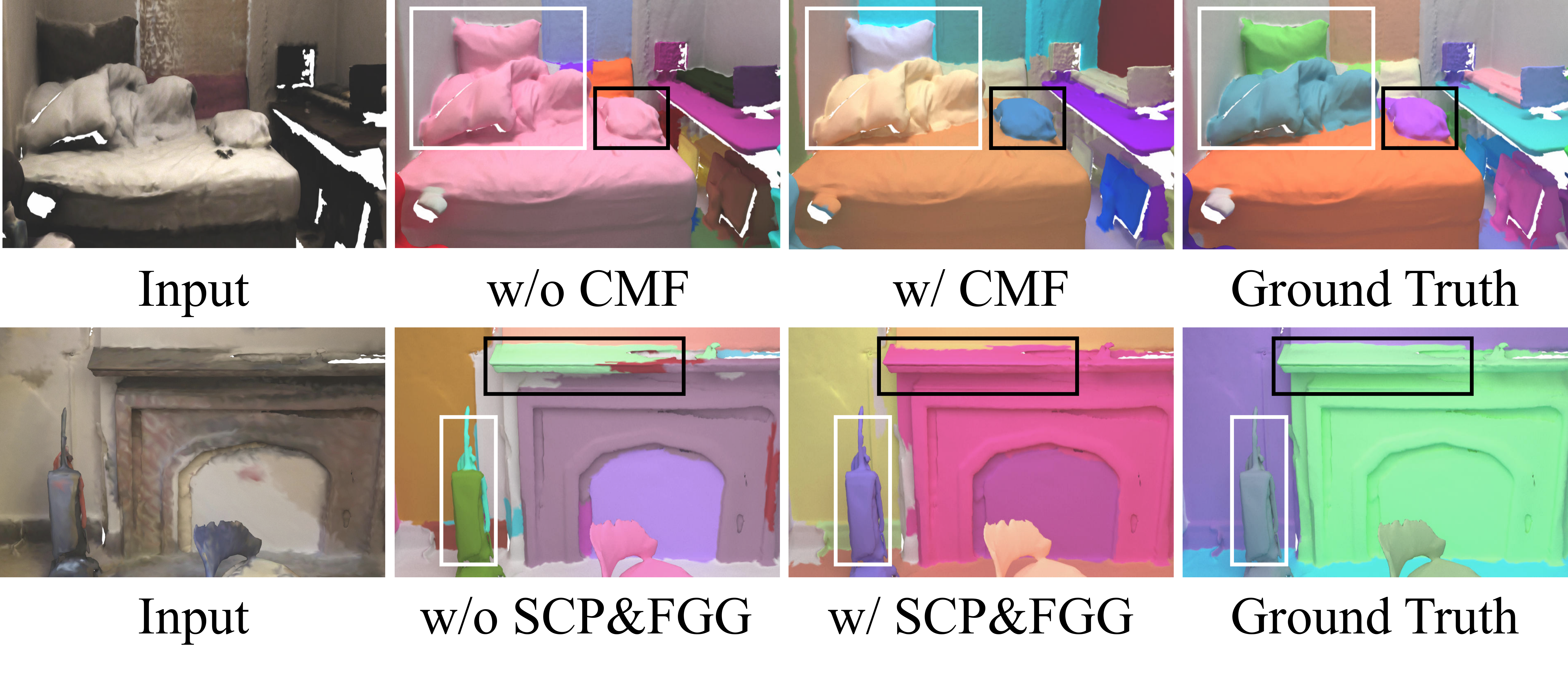}
    \caption{Ablation visualization results. Top: 2D mask proposal section; Bottom: semantic-guided aggregation section.}
    \label{fig:ablation_visualization}
\end{figure}

\subsection{Analysis Experiments}               
\subsubsection{Quantitative Results}

As shown in Table \ref{table:quantitative results}, SGS-3D achieves SOTA performance among training-free methods. On the challenging, depth-less KITTI-360 dataset, Ours$^{-}$ establishes a remarkable +16.4\% mAP lead over the next best competitor, SAI3D, validating our method's robustness in outdoor scenes. On indoor benchmarks, Ours$^{-}$ also consistently surpasses prior SOTA methods on ScanNet200 and ScanNet++. Furthermore, with GT depth, Ours$^{+}$ sets a new performance ceiling (e.g., 34.3\% mAP on ScanNet200), outperforming all training-free and even several training-dependent approaches, showcasing our framework's strong potential.

\begin{table*}[!t]
    \centering
    \setlength{\tabcolsep}{7pt}
    \begin{tabular}{l|c|c|cccccc}
        \toprule
        Methods & Proc. \& Year & 3D Mask & mAP & AP$_{50}$ & AP$_{25}$ & mAP$_{head}$ & mAP$_{common}$ & mAP$_{tail}$ \\
        \midrule
        OpenMask3D & NeurIPS23 & Supervised & 15.4 & 19.9 & 23.1 & 17.1 & 14.1 & 14.9 \\
        OpenIns3D & ECCV24 & Supervised & 15.9 & 20.6 & 23.3 & 19.2 & 14.2 & 14.2 \\
        \midrule
        OVIR-3D & CoRL23 & Zero-shot & 9.3 & 18.7 & \underline{25.0} & - & - &- \\
        SAM3D & ICCVW23 & Zero-shot & 9.8 & 15.2 & 20.7 & 9.2 & 8.3 & 12.3 \\
        SAI3D & CVPR24 & Zero-shot & 12.7 & 18.8 & 24.1 & \underline{12.1} & \underline{10.4} & \underline{16.2} \\
        SAM2Object & CVPR25 & Zero-shot & \underline{13.3} & \underline{19.0} & 23.8 & - & - & - \\
        \textbf{Ours} & - & Zero-shot 
        & \textbf{21.1}$\,_\text{{+7.8}}$ 
        & \textbf{29.4}$\,_\text{{+10.4}}$  
        & \textbf{35.0}$\,_\text{{+10.0}}$ 
        & \textbf{21.6}$\,_\text{{+9.5}}$  
        & \textbf{20.0}$\,_\text{{+9.6}}$  
        & \textbf{22.0}$\,_\text{{+5.8}}$  \\
        \bottomrule
    \end{tabular}
    \caption{Semantic results on ScanNet200. Head, common, and tail categories represent object classes with high, medium, and low frequencies in the dataset, respectively. OpenMask3D \cite{takmaz2023openmask3d} and OpenIns3D \cite{huang2024openins3d} requires supervised training on ScanNet200. In fully supervised and zero-shot setting, our method surpass previous SOTA methods.}
    \label{tab:semantic}
\end{table*}

\begin{table}[htbp]
  \centering
  \setlength{\tabcolsep}{2pt}
  \resizebox{\columnwidth}{!}{
    \begin{tabular}{l|cccccc}
      \toprule
      Method & Img. Prop. & mAP & AP$_{50}$ & AP$_{75}$ & Time & Inst. Pred. \\
      \midrule
      Open3DIS & 10\% & 29.7 & 45.2 & 56.8 & 9.18s & 102 \\
      \midrule
      \multirow{3}{*}{Ours} & 10\% & 34.3 & 51.3 & 64.6 & 9.51s & 62 \\
           & 5\% & 33.7 & 51.1 & 64.5 & 4.94s & 66 \\
           & 2.5\% & 31.8 & 48.8 & 63.2 & 2.42s & 68 \\
      \bottomrule
    \end{tabular}
  }
  \caption{Influence of input image proportion on segmentation efficiency and performance. Inst. Pred. indicates predicted instance counts per scene.}
  \label{table:efficiency}
\end{table}

\begin{table}[htbp]
  \centering
  \begin{minipage}[t]{0.45\linewidth}
    \centering
    \setlength{\tabcolsep}{2pt}
    \begin{tabular}{l|ccc}
      \toprule
      VFM & mAP & AP$_{50}$ & AP$_{25}$ \\
      \midrule
      Cropformer & \underline{33.6} & \underline{50.7} & 63.1 \\
      SAM        & 32.3            & 47.8             & \underline{63.8} \\
      YoloW-SAM  & 32.4            & 49.5             & 62.3 \\
      GD-SAM     & \textbf{34.3}   & \textbf{51.3}    & \textbf{64.6} \\
      \bottomrule
    \end{tabular}
    \caption{Ablation of different 2D vision models.}
    \label{table:2d_model}
  \end{minipage}
  \hfill
  \begin{minipage}[t]{0.4\linewidth}
    \centering
    \setlength{\tabcolsep}{3pt}
    \begin{tabular}{l|ccc}
      \toprule
      $c_{m}$ & mAP & AP$_{50}$ & AP$_{25}$ \\
      \midrule
      0.4 & 33.8 & 50.2 & 63.1 \\
      0.3 & 34.0 & 50.9 & 63.2 \\
      0.2 & 34.3 & 51.6 & 64.6 \\
      0.1 & 32.0 & 48.3 & 61.8 \\
      \bottomrule
    \end{tabular}
    \caption{Ablation of co-occurrence score $c_{m}$.}
    \label{table:co-occurrence score}
  \end{minipage}
\end{table}

\subsubsection{Qualitative Results}
Figure \ref{fig:qulitive_scannet_kitti} shows our method produces visually superior results with more complete and geometrically precise instances. This stems from our framework's ability to mitigate semantic ambiguity through mask filtering and spatial splitting, yielding cleaner and more coherent segmentations.

\subsubsection{Ablation Studies}
Table \ref{table:ablation} details the contribution of each component. CMF provides an initial boost. Critically, adding SCP without growing achieves 30.1\% mAP, already surpassing SOTA methods like SAM-graph and proving the power of our semantic purification. Finally, integrating FGG elevates performance to our full model's 34.3\% mAP, confirming a powerful synergy where our purification stages create high-quality seeds for FGG to complete (Figure \ref{fig:ablation_visualization}).

\subsubsection{Efficiency and Robustness}
Table \ref{table:efficiency} highlights SGS-3D's exceptional efficiency. Our method achieves superior accuracy to Open3DIS while using only 2.5\% of the images (vs. 10\%) and being nearly four times faster. It also significantly reduces over-segmentation by ~40\%.

Our pipeline demonstrates robustness across diverse vision foundation models (VFMs, Table \ref{table:2d_model}) \cite{qi2022high, cheng2024yolo} and is insensitive to the co-occurrence threshold $c_{m}$ within a reasonable range (Table \ref{table:co-occurrence score}). As shown in Table \ref{table:occlusion}, using a protocol inspired by \cite{naseer2021intriguing}, our method is also resilient to occlusion, maintaining strong performance even when up to 50\% of foreground masks are removed.

\begin{table}[htbp]
  \centering
  \setlength{\tabcolsep}{2pt}
  \begin{tabular}{c|cccccccc}
    \toprule
    Percentage (\%) & 0 & 5 & 10 & 30 & 50 & 60 & 70 & 90 \\
    \midrule
    mAP & 34.3 & 34.1 & 34.1 & 33.4 & 29.6 & 24.2 & 14.8 & 0.1 \\
    \bottomrule
  \end{tabular}
  \caption{Robustness to simulated occlusion.}
  \label{table:occlusion}
\end{table}

\subsection{Open-Set Scene Understanding Application}
\begin{figure}[t]
    \centering
    \includegraphics[width=0.45\textwidth]{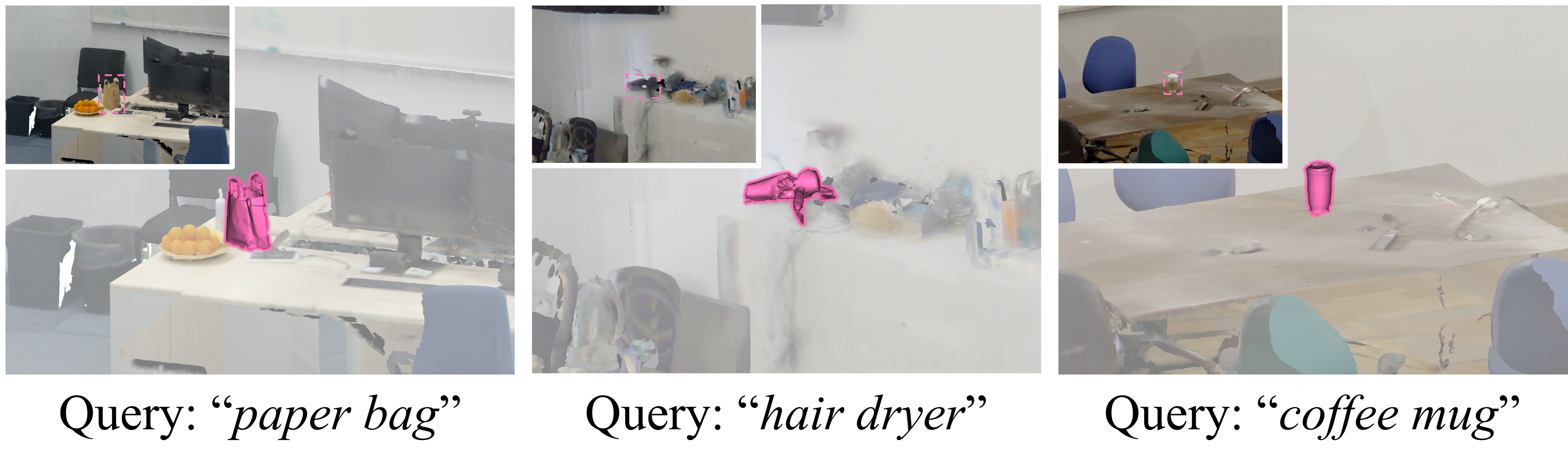}
    \caption{Open-vocabulary 3D object search. Given a text prompt, our method can accurately locate the corresponding object instances (pink) in a 3D scene.}
    \label{fig:semantic}
\end{figure}

Our high-quality, class-agnostic proposals form a strong basis for open-set understanding. By labeling them with the off-the-shelf vision-language models \cite{radford2021learning, zhai2023sigmoid}, we extend our method to open-vocabulary 3D segmentation. As shown in Table \ref{tab:semantic}, our geometrically precise instance masks reduce semantic ambiguity, enabling SGS-3D to significantly outperform previous zero-shot methods and support applications like text-based object search within complex 3D environments, as illustrated in Figure \ref{fig:semantic}.

\section{Conclusion}
We have presented SGS-3D, a training-free framework for high-fidelity 3D instance segmentation. Our "split-then-grow" strategy first purifies noisy proposals into high-quality 3D seeds using geometric and spatial cues, then grows them into complete instances via a feature-guided process. Experiments show SGS-3D substantially outperforms SOTA methods, particularly on challenging depth-less dataset. Our work provides a robust bridge between 2D semantics and 3D geometry, advancing class-agnostic segmentation and open-set understanding. 
Future directions include extending our framework to dynamic scenes and optimizing its multi-stage design for real-time performance.

\appendix
\section{Acknowledgments}
This work was supported in part by the National Natural Science Foundation of China under Grant 42371343, and in part by the Basic and Applied Basic Research Foundation of Guangdong Province under Grant 2024A1515010986.


\bibliography{aaai2026}

\end{document}